\documentclass[conference]{IEEEtran}
\IEEEoverridecommandlockouts
\usepackage{cite}
\usepackage{stfloats} 
\usepackage{amsmath,amssymb,amsfonts}
\usepackage{algorithmic}
\usepackage{graphicx}
\usepackage{textcomp}
\usepackage{xcolor}

\def\BibTeX{{\rm B\kern-.05em{\sc i\kern-.025em b}\kern-.08em
    T\kern-.1667em\lower.7ex\hbox{E}\kern-.125emX}}
\begin{document}

\title{The combination of context information to enhance simple question answering\\
}

\author{\IEEEauthorblockN{1\textsuperscript{st} Zhaohui Chao}
\IEEEauthorblockA{\textit{School of Computer Science and Technology} \\
\textit{Wuhan University of Technology}\\
Wuhan, China \\
chaozhaohui@whut.edu.cn}
\and
\IEEEauthorblockN{2\textsuperscript{nd} Lin Li}
\IEEEauthorblockA{\textit{School of Computer Science and Technology} \\
\textit{Wuhan University of Technology}\\
Wuhan, China \\
cathylilin@whut.edu.cn}
}

\maketitle

\begin{abstract}
With the rapid development of knowledge base, question answering based on knowledge base has been a hot research issue. In this paper, we focus on answering single-relation factoid questions based on knowledge base. We build a question answering system and study the effect of context information on fact selection, such as entity's notable type, out-degree. Experimental results show that context information can improve the result of simple question answering.
\end{abstract}

\begin{IEEEkeywords}
question answering, knowledge base, context information
\end{IEEEkeywords}

\section{Introduction}
Question answering (QA) is a classic natural language processing task, which aims at building systems that automatically answer questions formulated in natural language\cite{b1}.

In recent years, several large-scale general purpose knowledge bases (KBs) have been constructed, including Freebase\cite{b2}, YAGO\cite{b3}, DBpedia\cite{b4} and Wikidata\cite{b5} . In addition, there are some commercial KBs that are not completely open, such as Google Knowledge Graph  and the Facebook Graph . We can access the data of entities and relationships through a specific API. Since Google put forward the concept of knowledge graph, the related research of KBs has reached a new level of popularity. 

Most of KBs store information in the form of RDF triples (subject, predicate, object)\cite{b6,b7}. For example, (/m/02mjmr, /people/person.place\_of\_birth,  /m/02hrh0\_), where \textbf{/m/02mjmr} is the Freebase id for Barack Obama, and \textbf{/02hrh0\_} is id for Honolulu. By structuring knowledge storage in this basic form, we can better organize, manage, and utilize vast amounts of knowledge. But people cannot directly understand and extract the knowledge in the knowledge graph without struct query language. So by asking questions, mapping the questions to the triples in the knowledge base and getting the correct answers, this is a good way to use the knowledge base. With the development of KBs, knowledge base-based question answering (KB-QA) has attracted more and more researcher's attention.

KB-QA is defined as the task of retrieving the correct entity or set of entities from a KB given a query expressed as a question in natural language\cite{b1}. For instant, in order to answer the question ``where was former U.S. President Obama born? " we need to retrieve the entity \textbf{/m/02mjmr} in Freebase to represent former U.S. president Barack Obama and the relation \textbf{/people/person.place\_of\_birth} corresponds to this question. With this entity and relation, form a corresponding structured query, and then obtain the right answer \textbf{Honolulu} (id: /m/02hrh0\_). Typically, with SPARQL (an RDF query language) people can extract information they needed from KB.

For each triple (entity1, relation, entity2), it shows the relationship between entity1 and entity2. The relation consists of entity's type and properties, defined by the Freebase ontology\cite{b6}.  
In this paper, we focus on answering single-relation factoid questions (simple questions), which can be answered by a single fact of the KBs. It is also called simple question answering (Simple QA). The aim of this paper is to analyze the effect of context information to answer selection. Combining entity's out-degree and notable type improves the accuracy of answer. In summary, our main works are as follows:

(a) Construct models for entity recognition and relationship matching. 

(b) In relation detection, we propose a simple and efficient method for constructing training datasets. Combined with a simple network structure, it can achieve a performance on relation detection.

(c) Explore the effects of entity's out-degree, notable type to the accuracy of simple question answering.
 
\begin{figure*}[ht]
\centerline{\includegraphics[scale=0.75]{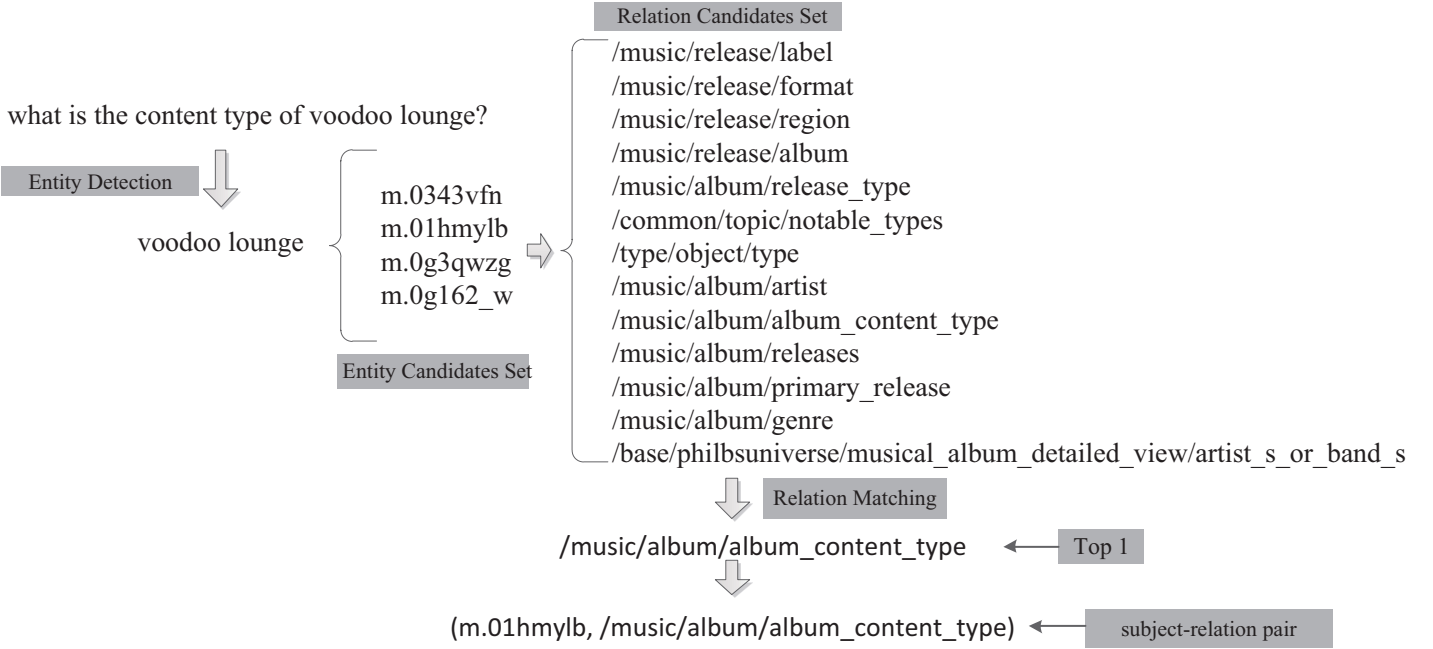}}
\caption{Example of Simple QA.}
\label{fig}

\end{figure*}

\section{Related work}
The core idea of deep learning system in KB-QA tasks is representation and matching. First, learn representations of both the question and the fact of KB, which contains literature level and semantic level. Then calculate the correlation between question and fact.

The work of answering single-relation factoid questions was first proposed by Bordes et al.\cite{b8}. In this work, they employ Memory Networks and introduce a new dataset SimpleQuestions, which was built on Freebase.

Bordes et al.\cite{b1,b9} use embedding models to map the question and knowledge to same space and matrix the similarity of their presentations.Many researchers have explored how to apply attention mechanism in image recognition. Wu et al. \cite{b10,b11,b12} employ attention mechanisms and joint learning. And they design an end-to-end network structure. Golub and He\cite{b13} employ a character-level, attention-based encoder-decoder framework for question answering. The model is robust for unseen entities since it adopts character-level modeling. Dai et al.\cite{b14} propose Conditional Focused neural network. Yin et al.\cite{b15} utilize a character-level convolutional network (char-CNN) to match subject entity in the question. In the task of matching the relation in the question, they use a word-level CNN (word-CNN). 

Lukovnikov et al.\cite{b16} present an end-to-end neural network. They merge word- and character-level representations of question. Hao et al.\cite{b17} present an end-to-end neural network model to represent the questions, which improves the representation of questions via cross-attention mechanism. Qu et al.\cite{b18} propose an attentive recurrent neural network with similarity matrix based convolutional neural network model, which combine the advantages of RNN and CNN.

\section{Approach}

\subsection{Task definition}
Single-relation factoid questions (simple questions) can be answered by a single fact of the KBs.
The formal definition is as follows. Knowledge base \( \{s_i, r_i,o_i\} \)  is a set of triples, where \( s_i \) and  \( o_i \)  are the subject entity and object entity,  \( r_i \)  is the relation,  \( (s_i, r_i,o_i) \)  corresponds to one fact. For question \textbf{q} formulated in natural language, find a triple , where \( s \)   and \( o \)  correspond to the subject and predicate in the question \textbf{q},  \( o \) is the answer to question \textbf{q}. 

So as long as we find the corresponding subject and predicate, we can turn question into a structured query to obtain the answer. 

\subsection{Model description}
We divide the task into two subtasks: entity detection and relation detection. First generate entity candidates set \textbf{E} by entity detection. 

Based on entity candidatess set \textbf{E}, we obtain all of the relations associated with the entity candidates. Then we calculate the semantic similarity between the relation and the question by semantic matching model. We take the relation \textbf{r} with the highest matching score as the answer to relation detection. Finally we select the corresponding triple as the answer based on \textbf{r}. As shown in the picture 1, after getting entity candidates set,``/music/album/album\_content\_type" is the highest matching relation with the question. At this time, we get the subject-relation pair. According the subject-relation pair, we can find the triple (m.01hmylb, /music/album/album\_content\_type, m.06vw6v) as the fact to answer the question.

\subsubsection{Entity Detection}
Following the traditional approach, we first conduct entity recognition. Mark out the entity text in the question, the words that belong to the entity. Like the mainstream method, we treat it as a sequence labeling task. We train a bidirectional LSTM\cite{b19,b20} network to detect entity text in the question.

\begin{figure}[htbp]
\centerline{\includegraphics[width=0.49\textwidth]{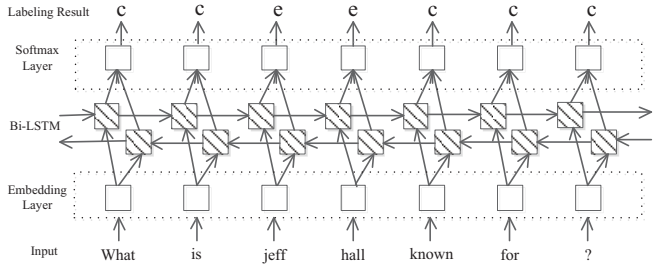}}
\caption{The architecture of entity recognition: use Bi-LSTM for entity recognition to extract the entity text.}
\label{fig}
\end{figure}

As shown in Figure 2, the words that belong to the entity text are marked \textbf{e}, and the words that do not belong to the entity text are marked \textbf{c}. The fragment of entity text refers to the sequence of words corresponding to each consecutive segment \textbf{e}.

Because some result of entity recognition may not completely correct, we have employ some strategies to remedy it. In order to solve this problems, we borrow the method proposed by Ture and Jojic\cite{b21}. Based on the result of entity recognition, we obtain the fragment of entity text. Then we construct entity candidates sets through the following process.

(a) Find out all entities in FB2M whose alias exactly equal to the fragment of entity text. Then form entity candidates set \textbf{E}. If there is no exact match entity, go to the second step.

(b)  Extract 1-gram, 2-gram and 3-gram from each fragment of entity text. If a tuple is a subset of another tuple, keep the long one and discard the short one. Then form a set of n-gram. We can search the entities based n-gram and form a set of entity candidates S. Using formula (1) to calculate the weight of the entity. Then we add the entity with the weight equal to the highest score to the candidates set \textbf{E}.

\begin{equation}
 \begin{split}
score_i= \frac {N_i}{ L_iC_i} 
\end{split}
\end{equation}

Where \( N_i \)  is the number of words in n-gram \( G \) , \( L_i \) is the number of words in the entity's name , \( C_i \) is the number of entities retrieved based on \( G \) .

\subsubsection{Relation detection}
In this subtask, our core mission is to measure the semantic similarity between the relations and questions. Therefore, we design a semantic matching model.

Network structure: We take the matching task as a binary classification problem, matching or not matching. First we design a binary classification network. Instead of using the classification result directly, we use the value of the output layer as the matching score.

As shown in picture 3, in this progress, we use bidirectional GRU\cite{b22} network encode the question and relation to generate their sematic representation. We then concatenate the two vector representation and feed the concatenated vector into a dense layer to produce the final classification result. We use sigmoid as the activation function of output layer.

\begin{figure}[htbp]
\centerline{\includegraphics[width=0.5\textwidth]{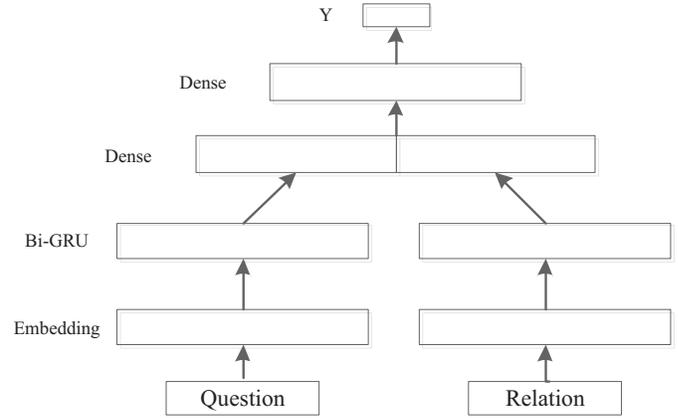}}
\caption{Network structure of semantic matching model.}
\label{fig}
\end{figure}

Generation training data: According to the first word of each relation, we divide the relation into 89 major categories, representing 89 domains. For example, in the domain of music, several relation examples are shown in table 1. 


\begin{table}[htbp]
\caption{Several relation examples in music domain}
\renewcommand\arraystretch{1.25}
\begin{center}
\begin{tabular}{|c|}
\hline
\textbf{/music/live\_album/concert\_tour} \\
\textbf{/music/composition/compose} \\
\textbf{/music/release/label} \\
\textbf{/music/album\_content\_type/albums} \\
\textbf{/music/recording/producer} \\
\textbf{/music/genre/parent\_genre} \\
\textbf{/music/artist/concert\_tours} \\
\textbf{/music/album/compositions} \\
\textbf{......} \\
\hline
\end{tabular}
\label{tab1}
\end{center}
\end{table}

We generate training data in units of questions. For each question q and the triple  \( (s, r,o) \), we first determine the domain \textbf{D} according to the golden relation \textbf{r}. Then we get all the relations \textbf{R} of the domain \textbf{D}. Next we generate pairs formatted in the form of  \( (q, R_i,tag) \) where tag is equal to 0 or 1. As illustrated in picture 4, the relation corresponding to the question belongs to the music domain. Thus, we pair all the relations belonging to the music domain with questions and form corresponding tags.

\begin{figure}[htbp]
\centerline{\includegraphics[width=0.5\textwidth]{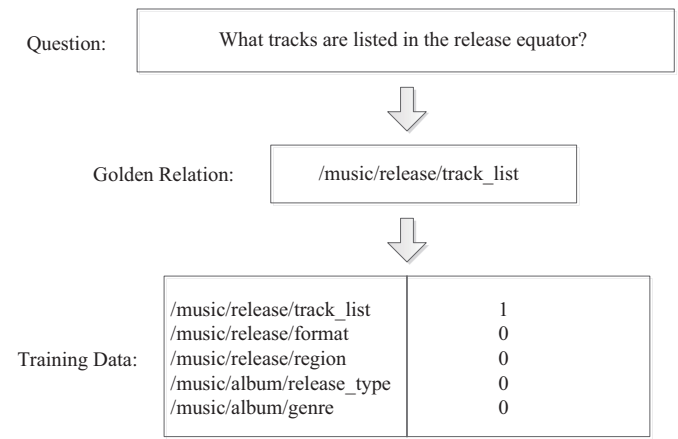}}
\caption{Example of generating relation matching training data.}
\label{fig}
\end{figure}

In addition, we copy positive cases three times in order to reduce or avoid the effects of data imbalances. Because in the construction of training data, the proportion of negative samples generated is much larger than positive samples.

\subsection{Context information}
In this paper, we explore to improve the resolution of the entities with the same name by entity's context information. Here context information includes entity's out-degree and notable type.

\subsubsection{out-degree of entity}
The out-degree of entity e is the number of triples in the KG in which e is the subject. We try to rank the entity candidates set based on the out-degree.
\subsubsection{Notable types}
FreeBase's notable types are simple atomic labels given to entities that indicate what the entity is notable for\cite{b23}.  Freebase was acquired by Google In 2010 and officially shut down in 2016. Its data was migrated to Wikidata. Since Freebase's online API is closed, it is not possible to get the notable type information directly. The Freebase data dumps  can be downloaded in an N-Triples RDF format.

We extract the notable type information from the dump files. There are 1275 kind of notable types in 2 million entities of FB2M. We try to use that the type information of the entity to distinguish the entities with same name. 

We continue to use the same network structure like relation detection to calculate the matching score between question and notable type. And we try to improve the accuracy of entity recognition by ranking candidates entities based on the matching score.


\begin{table}[htbp]
\caption{The accuracy of entity recognition.}
\renewcommand\arraystretch{1.5}
\begin{center}
\begin{tabular}{|p{2.5cm}|p{2.5cm}|}
\hline
Item	  & Percentage  \\
\hline
unique  &	50.5\%   \\
\hline
Not unique  & 31.7\%  \\
\hline
Total  & 82.2\%  \\
\hline
\end{tabular}
\label{tab1}
\end{center}
\end{table}


\begin{table}[htbp]
\caption{Compare the use of relation, notable type and relation for results of entity selection.}
\renewcommand\arraystretch{1.5}
\begin{center}
\begin{tabular}{|c|c|c|}
\hline
Method & Error with same name entity & Accuracy \\
\hline
base & 	15.3\% &	65.5\% \\
\hline
base + out-degree  &	14.2\%  &	66.6\% \\
\hline
base + subject type	 & 14.6\%	 & 66.2\%  \\
\hline
baseline (Bordes et al.\cite{b8})  &	 -  &		62.7\% \\
\hline
\end{tabular}
\label{tab1}
\end{center}
\end{table}


\section{Experiments}
\subsection{Dataset}
Data Set: we utilize SimpleQuestion dataset, released by Bordes et al.\cite{b8}. It consists of 108442 questions written in natural language. It is constructed according corresponding facts in Freebase. The facts format as (subject, relationship, object). According to the original data division ratio, there are 75910 training data, 21687 test data, and 10845 validation data. This dataset also provides two subsets of Freebase: FB2M and FB5M. They are represented as a set of triples. We take FB2M as background knowledge base, includes 2 million entities and 6701 relations. In addition, we also use Freebase data dumps (22 GB compressed, 250 GB uncompressed) to extract entity's notable type information.

\subsection{Training settings}
The model word embeddings are initialized with the 300-dimensional pre-trained vectors provided by Glove\cite{b24}. We update network weights by using the Adam\cite{b25} optimizer with learning rate 0.001. The hidden layers of Bi-LSTM and Bi-GRU have size 100. In the semantic matching model, Dropout is set to 0.1.
\begin{table}[htbp]
\caption{Accuracy on QA accuracy. ``base" means we don't use the context information to distinguish entities with same name.}
\renewcommand\arraystretch{1.5}
\begin{center}
\begin{tabular}{|c|c|c|}
\hline
Method & Error with same name entity &  Accuracy \\
\hline
(a)no rank	 &  26.1\%  & 	59.5\% \\
\hline
(b)notable type	 &  24.5\%  & 	60.2\%  \\
\hline
(c)out-degree	 &  15.4\%  & 	64.0\%  \\
\hline
\end{tabular}
\label{tab1}
\end{center}
\end{table}


\begin{table}[htbp]
\caption{The recall of top K entity candidatess. ``Ranking" means ranking the entity candidates set based on the out-degree.}
\renewcommand\arraystretch{1.5}
\begin{center}
\begin{tabular}{|p{1.5cm}|p{1.5cm}|p{1.5cm}|}
\hline
Top K	  & Not ranking  &	Ranking  \\
\hline
1  &	60.5\% &	70.5\%  \\
\hline
5  & 74.1\% &	80.9\% \\
\hline
10 &	77.0\% &	82.6\% \\
\hline
400 &	86.4\% & 	87.4\%  \\
\hline
\end{tabular}
\label{tab1}
\end{center}
\end{table}

\subsection{Initial preparation work}
(1)Divide allentitiy names into words and generation 1-gram, 2-gram and 3-gram. Then build an inverted index \textbf{I} that maps all n-grams to the entity's alias text.

(2) Extract entity's notable type information from Freebase data dumps.

(3) According to training set, labeling the question text and generating an entity recognition training set.

\subsection{Evaluation}
Only if the subject and the relation are correctly predicted, the question \textbf{q} is considered being answered correctly. So we use accuracy of subject-relation pair to measure the final question answering results.

\subsection{Experimental results and discussion}

\subsubsection{Entity Detection}
Table 2 shows the result of entity recognition. The accuracy rate is 82. 2\%, of which 31.7\% of entities cannot be uniquely identified. Because one name or alias may corresponds to multiple entities. In this case, the corresponding entity cannot be uniquely identified by the name or alias alone. 
Besides, 17.8\% of entities are not fully labeled correctly.This part needs to retrieve entities based on the result of entity recognition and form entity candidates sets .

\subsubsection{Relation Detection}
We generate a test data set for the model of relation detection based on the test set. For each question, 
we take the relations associated with its golden entity as the relation candidates set. Then use the model to find the highest matching relation.
After testing, accuracy rate of relation matching can reach 90.1\%.

\subsubsection{Combination with out-degree and notable type information}
As shown in table 3,our base model achieves accuracy of 65.5\% on FB2M. Combined with out-degree, the final accuracy can be increased by 1.1\%. After add notable information, the final result can also be increased by 0.7\%. This demonstrates combining context information can improve the model's ability to distinguish entities of the same name.

In the experiment, we also try to directly take the top 1 entity of candidates set \textbf{E} as the answer of entity detection. We try the following three methods and compare the result .

(a) Don't rank the entities by any context information. We directly use the top 1 entity e. And all relations associated with e are added to the relation candidates set R. Then we use the semantic matching model to select the relation r with the highest matching score as the answer of relation detection. Therefore, the subject-relation pair (e, r) is the final answer. According this method, the accuracy of final result achieves 59.5\%.

(b)Rank the entities by the semantic matching score between the notable type information and question. Next, like the method, we get the final answer. The accuracy becomes 60.2\%.

(c)The difference from method (b) is that we use the out-degree to rank the entities. The final accuracy increases to 64.0\%, which is only lower 1.5\% than the result of the base model in table 2.

All the results are shown in table 4. For the SimpleQuestion dataset, the entity's out-degree can effectively improve the resolution of the entities with the same name. 
Through the influence of out-degree, we can see that the frequency can help us to effectively distinguish the entity with the same name. Entity with more out-degree indicates that the entity has more contact with other entities. So in comparison, such entities are more likely to be correct answers.

Besides, as shown in table 5, combined with the context information, the recall of Top-K can be greatly improved. So with these context information, the candidates space of the entity can be effectively reduced and the retrieval efficiency can be improved.

\section{Conclusions}

In this paper, we compare the impact of entity's out-degree, notable type information on answering single-relation factoid questions based on KBs. 
Combining context information can improve the accuracy of answering. It helps to distinguish ambiguity of the entity with the same name. In addition, we find there are some ambiguities that cannot be resolved in limited context information. In practical applications, the combination of the questioner's identity information (user profile information) and more context information may solve the problem to some extent. In future work, we will explore how to efficiently generate entity candidates set and deal with multi-relation problem.

\vspace{12pt}

\end{document}